\documentclass[10pt,twocolumn,letterpaper]{article}

\usepackage{cvpr}              % To produce the CAMERA-READY version

\usepackage{multirow}
\usepackage{booktabs}
\usepackage{pifont}

\usepackage[dvipsnames]{xcolor}
\definecolor{cvprblue}{rgb}{0.21,0.49,0.74}
\usepackage[pagebackref,breaklinks,colorlinks,citecolor=cvprblue]{hyperref}

\def\paperID{3997} % *** Enter the Paper ID here
\def\confName{CVPR}
\def\confYear{2024}

\title{Rectify the Regression Bias in Long-Tailed Object Detection}

\author{Ke Zhu \quad Minghao Fu \quad Jie Shao \quad Tianyu Liu \quad Jianxin Wu\thanks{J. Wu is the corresponding author.}\\
State Key Laboratory for Novel Software Technology, Nanjing University, China \\
School of Artificial Intelligence, Nanjing University, China \\
{\tt\small \{zhuk,fumh,shaoj,liuty\}@lamda.nju.edu.cn, wujx2001@nju.edu.cn}
}

\begin{document}
\maketitle

\begin{abstract}
    Long-tailed object detection faces great challenges because of its extremely imbalanced class distribution. Recent methods mainly focus on the classification bias and its loss function design, while ignoring the subtle influence of the regression branch. This paper shows that the regression bias exists and does adversely and seriously impact the detection accuracy. While existing methods fail to handle the regression bias, the class-specific regression head for rare classes is hypothesized to be the main cause of it in this paper. As a result, three kinds of viable solutions to cater for the rare categories are proposed, including adding a class-agnostic branch, clustering heads and merging heads. The proposed methods brings in consistent and significant improvements over existing long-tailed detection methods, especially in rare and common classes. The proposed method achieves state-of-the-art performance in the large vocabulary LVIS dataset with different backbones and architectures. It generalizes well to more difficult evaluation metrics, relatively balanced datasets, and the mask branch. This is the first attempt to reveal and explore rectifying of the regression bias in long-tailed object detection.
\end{abstract}

\section{Introduction}
\label{sec:intro}

Long-tailed object detection draws great attention~\cite{LVIS} for its practical utility recently. Numerous efforts~\cite{LTD_EQL,LTD_ACSL,LTD_ECM,LTD_SeeSaw,LTD_GumbelLoss} have been made to tackle this challenging task, such as re-weighting~\cite{LTD_EQLv2,LTD_ACSL,LTD_SeeSaw}, over-sampling~\cite{LVIS,LTD_LOCE,LTD_ForestRCNN}, and balanced grouping~\cite{LTD_LST, LTD_BAGS}. These methods are proposed to prevent the tail classes from being overwhelmed due to discouraging gradients~\cite{LTD_EQLv2,LTD_ACSL}, inferior predicting scores~\cite{LTD_LOCE,LTD_BAGS} or insufficient samples~\cite{Det_LibraRCNN,LTD_LOCE}.

Long-tailed detection often involves both classification and regression branches. While almost all existing methods focus on mitigating the classification bias (\eg, adjusting the classification structure in detection branches), little or \emph{no attention has been paid to the regression branch}. We will show in this paper that the regression bias \emph{has significant adverse effects in long-tailed object detection}, but previous methods failed to identify this important issue.

\begin{figure}
	\centering
    \begin{subfigure}{0.49\linewidth}
		\includegraphics[width=0.95\linewidth]{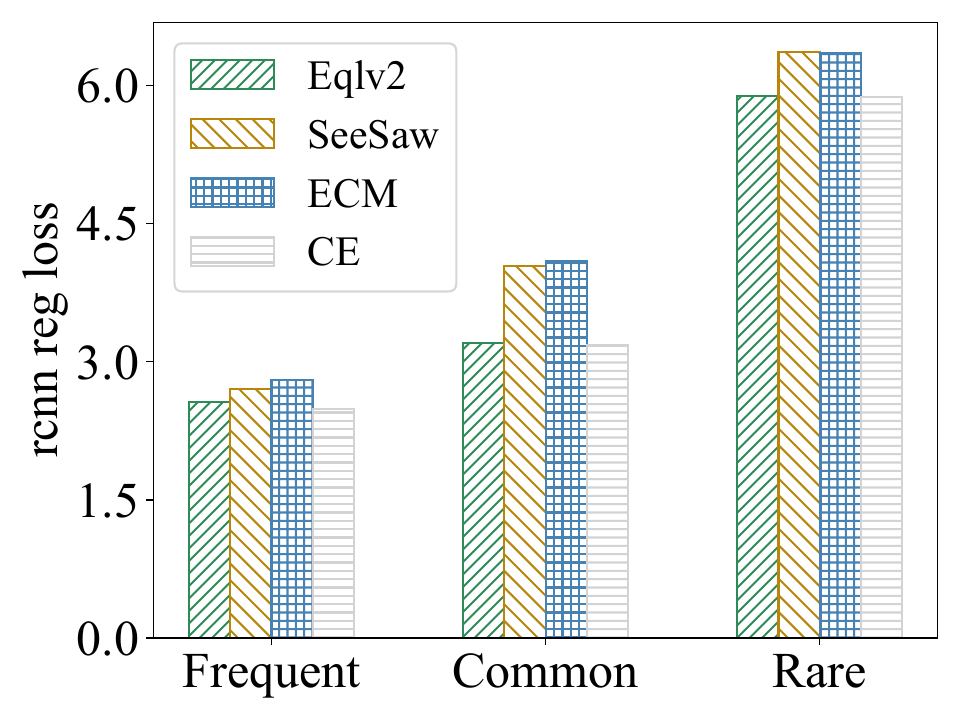}
		\caption{RCNN regression train loss}
		\label{fig:motivation:rcnn}
	\end{subfigure}
	\begin{subfigure}{0.49\linewidth}
		\includegraphics[width=0.95\linewidth]{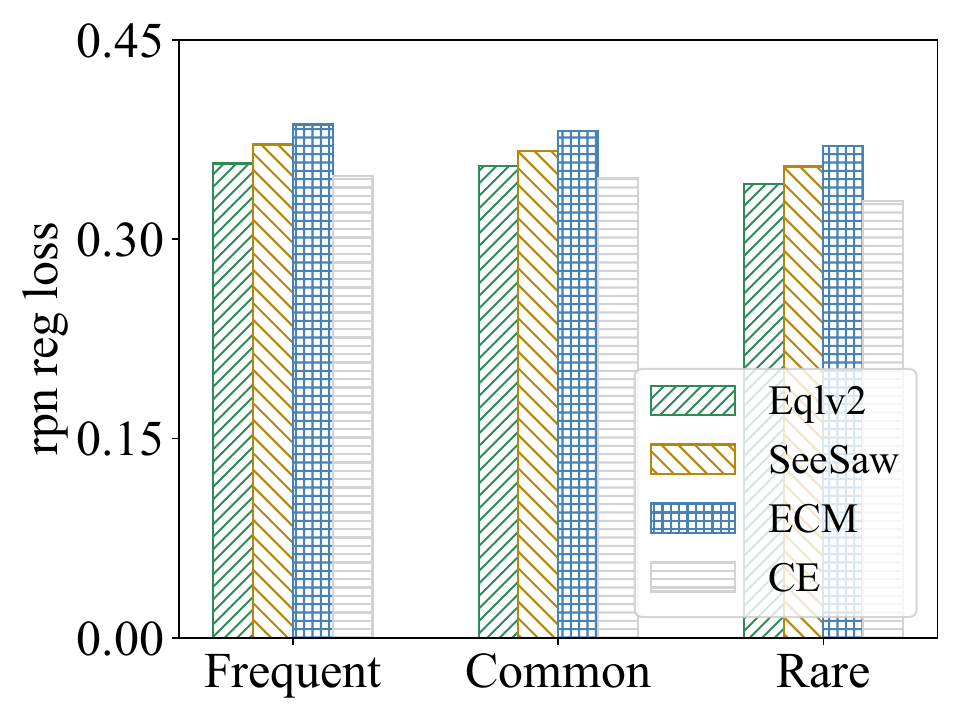}
    	\caption{RPN regression train loss}
		\label{fig:motivation:rpn}
    \end{subfigure} 

    \begin{subfigure}{0.95\linewidth}
		\includegraphics[width=1.0\linewidth]{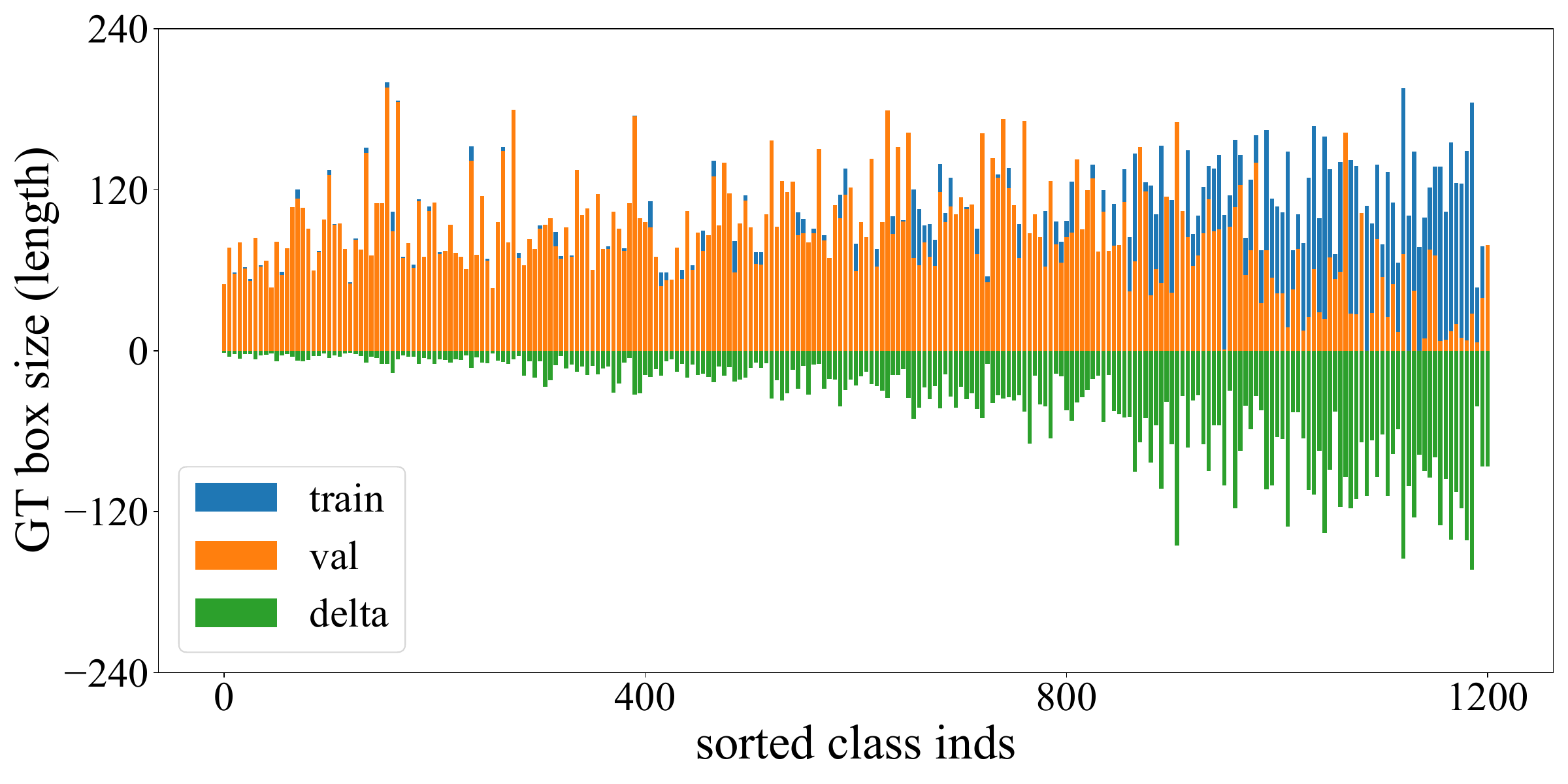}
    	\caption{Mean scale in LVIS1.0 train and validation set}
		\label{fig:motivation:scale}
    \end{subfigure}
	\caption{\ref{fig:motivation:rcnn} shows the RCNN regression loss of frequent, common and rare categories. \ref{fig:motivation:rpn} shows the RPN regression loss. \ref{fig:motivation:scale} shows distribution of per class mean object scales in LVIS1.0. `delta' in~\ref{fig:motivation:scale} is the \emph{negative} of difference between train and validation set sizes for different classes.}
	\label{fig:motivation}
\end{figure}

Fig.~\ref{fig:motivation} clearly showcases this issue. For different methods 
(EQLv2~\cite{LTD_EQLv2}, SeeSaw~\cite{LTD_SeeSaw}, ECM~\cite{LTD_ECM} and CrossEntropy CE~\cite{LVIS}) trained on the LVIS1.0~\cite{LVIS} dataset, we plot the regression branch's losses for the final detection RCNN head in Fig.~\ref{fig:motivation:rcnn}. It is clear that the regression losses of rare categories are significantly higher than those of frequent and common categories, which inevitably will lead to poor regression results (and hence detection results) for rare classes. We name this finding as the \emph{regression bias}, but existing long-tailed detection methods all fail to deal with or even identify the regression bias.

To further demonstrate the importance of regression, we calculate the class-wise mean scale of GT (groundtruth) boxes in LVIS train and validation sets, as well as their differences (\cf. Fig.~\ref{fig:motivation:scale}). The scale shift of rare classes is much larger than that of frequent classes. Since regression is highly correlated with box scale~\cite{faster-rcnn}, it is thus inherently difficult for a rare class to learn a good bounding box regressor with both few samples and large scale shift. In short, \emph{it is crucial to properly handle the regression bias} in long-tailed object detection.

Our solution to rectify the regression bias is motivated by Fig.~\ref{fig:motivation:rpn}. We find that the regression loss in RPN is \emph{balanced} where rare, common and frequent categories have almost the same regression loss, which is almost immune to the regression bias when comparing Fig.~\ref{fig:motivation:rpn} with Fig.~\ref{fig:motivation:rcnn}.

The key difference between RPN and RCNN regression is that the former is \emph{class-agnostic} (i.e., all classes share the same regression parameters), while the latter is \emph{class-specific}. Then, one natural question arises: Can class-agnostic head improve a tail class's generalization ability and handle the regression bias? Our hypothesis is that \emph{rare classes do favor class-agnostic regression heads}.

Our hypothesis is supported by experiments in Table~\ref{tab:motivation}, where we compared class-specific and -agnostic regression heads in RCNN. By replacing the classification head with groundtruth class labels, experiments in Table~\ref{tab:motivation} disentangled the impact of the classification head and focus on the regression heads. It is clear that the class-agnostic head possesses substantial advantages: AP$^b_r$ (for rare) surges from 0.7 to 54.6, even surpassing AP$^b_f$ (for frequent classes)! However, the agnostic head will bring a small drop in frequent classes (\eg, from 40.7 to 40.0 in AP$_{f}^b$). \emph{The final version of our conjecture} is: the rare (and possibly common) classes indeed favor class-agnostic regression, while the frequent classes prefers a class-specific regression, and there should be a \emph{trade-off} between the two to optimize for all three types of categories.

\begin{table}
	\setlength{\tabcolsep}{5pt}
	\centering
	\begin{tabular}{lllllll}
		\toprule[1pt]
		 Reg. Head & GT & AP  &  AP$^{b}$ & AP$^{b}_{r}$ & AP$^{b}_{c}$ & AP$^{b}_{f}$  \\
		\midrule[1pt]
        \multirow{2}{*}{Specific} & \textit{no} & 18.7 &  19.7 & \phantom{0}0.3 & 16.5 & 31.7\\
         & \textit{yes} & \textbf{37.2} &  \textbf{39.8} & \textbf{34.1} & \textbf{41.3} & \textbf{40.7}\\
         \midrule
        \multirow{2}{*}{Agnostic} & \textit{no} & 18.0 &  19.3 & \phantom{0}0.7 & 15.5 & 31.5\\
         & \textit{yes} & \textbf{38.4} & \textbf{45.6} &  \textbf{54.6} & \textbf{47.0}  & \textbf{40.0}\\
		\bottomrule[1pt]
	\end{tabular}
\caption{Results on LVIS1.0 based on the CE baseline. We trained a Mask-RCNN R50-FPN detector~\cite{Mask-RCNN} with class-specific (the default setting) and class-agnostic regression heads in the RCNN head. Please note that during test, the predicted proposals were provided with groundtruth (GT) classification results.}
\label{tab:motivation}
\end{table}

Accordingly, we design three different methods to fully rectify the regression bias, including adding a class-agnostic head, clustering similar heads, or merging heads. All three methods bring in positive effects (\cf Table~\ref{tab:main-method}), which verifies that rectifying regression bias is indeed crucial. We choose to adopt `adding a class-agnostic head' in our main experiment for its simplicity, which leads to consistent and significant improvements over previous long-tailed detection pipeline, and has achieved state-of-the-art performance with various backbones and architectures. Moreover, our method shows robust generalization ability (\cf Tables~\ref{tab:other-metrics}-\ref{tab:mask-branch}) under varying settings, including different datasets (COCO/COCO-LT~\cite{LTD_ClassificationFirst}), different evaluation metrics and even adapts to the mask branch design. Furthermore, visualizations show that the proposed method indeed alleviates the regression bias (\cf Fig.~\ref{fig:flatten-distribution}) and leads to more accurate bounding box predictions (\cf Fig.~\ref{fig:box-refine}). In summary, our contributions are:
\begin{enumerate}
    \item For the first time, we reveal and successfully handle the regression bias in long-tailed object detection;
    \item We propose three remedies to alleviate this bias, all of which produce consistent improvements over existing methods.
    \item Our method achieves state-of-the-art results on LVIS as well as generalizing across datasets, metrics and even the mask branch. Visualizations qualitatively verify our hypothesis, too.
\end{enumerate}

\section{Related Work}
\label{sec:related-work}

Vast number of pipelines and techniques have been invented in the field of general purpose object detection. Although most of these detectors can possibly be adapted to balanced detection datasets~\cite{MS-COCO}, they can nevertheless hardly handle extremely long-tailed distribution. In this paper, we discover the regression bias in long-tailed object detection, and propose effective remedies to alleviate this bias.

Modern long-tailed learning methods aim to solve two important tasks: image classification~\cite{ImageNet} and object detection/segmentation~\cite{faster-rcnn,LVIS}.

The first task, long-tailed image classification~\cite{ImageNetLT}, is well explored and displays great diversity in terms of specific methods~\cite{LT_LDAM,BSCE}, training pipelines~\cite{LT_Decoupling,LT_BBN}, post-hoc analysis~\cite{LT_logit_adjustment,LT_logit_adjustgaussian} and network architecture~\cite{LT_KD_DiVE, LT_KD_SSL, LT_KD_RIDE}. Early attempts in this family focused on the details of re-weighting~\cite{LT_LDAM} or re-sampling~\cite{LT_Effective_Number} techniques to provide a relatively balanced (data) distribution for tail classes. Later on, the decoupling pipeline~\cite{LT_Decoupling} for long-tailed learning becomes popular, paving the way for various successors~\cite{LT_BBN, LT_Calibrate}. This decoupled paradigm are based on the insight that instance sampling is beneficial for representation learning, while the classifier needs to be post-calibrated to alleviate its bias. Some works try to unify long-tailed learning with theoretical analysis~\cite{LT_logit_adjustment, LT_logit_adjustgaussian}, which involves label distribution shift~\cite{BSCE} and generalization error bound~\cite{LT_logit_adjustment}. Most recent methods are much more diverse in ideas: they involve knowledge distillation~\cite{LT_KD_DiVE, LT_KD_RIDE, QFD,AAFM}, self-supervised learning techniques~\cite{LT_KD_SSL, MLS, Cropping} or statistical approach~\cite{Duyx} to better handle the long-tailed recognition task.

The combination of long-tailed learning and object detection or instance segmentation proves to be more challenging~\cite{LVIS}, because the class distribution in LVIS~\cite{LVIS} is extremely imbalanced and naively applying common object detection techniques leads to un-satisfactory results~\cite{LTD_EQL,LTD_BAGS,LTD_ACSL,LTD_DropLoss,LTD_SeeSaw}. Dominating solutions in this area are re-sampling~\cite{LVIS,LTD_ForestRCNN} and re-weighting~\cite{LTD_PCB,LTD_EQLv2,LTD_EQFL,LTD_SeeSaw,LTD_ACSL}. There are variants of them that adopt balanced grouping~\cite{LTD_BAGS}, class incremental learning~\cite{LTD_LST,ibm2}, augmented feature sampling~\cite{LTD_LOCE, LTD_ClassificationFirst} or extra data source~\cite{LTD_MosaicOS}. A common characteristic of these prior methods is that they only \emph{focus on the classification task, and ignore the subtle influence of the regression branch}. In this paper, we thoroughly explore the previously unvisited regression bias in long-tailed object detection, and propose a simple yet novel method to tackle long-tailed object detection.

In~\cite{LTD_ClassificationFirst}, there is an experiment with similar design as ours in Table~\ref{tab:motivation}. But, we emphasize that they are fundamentally different. On one hand, conclusions in~\cite{LTD_ClassificationFirst} is that `performance drop in LVIS is mainly caused by the proposal classification~\cite{LTD_ClassificationFirst}', and hence they focused on \emph{classification}. On the other hand, their class agnostic-head is a supplementary to support their classification hypothesis, while our design and experiments lead to a novel finding: the regression bias.

\section{Method}

Now we elaborate three different remedies to alleviate the regression bias, and start from the background on long-tailed object detection.

\subsection{Preliminaries}

We take Faster-RCNN~\cite{Mask-RCNN} as an example. For a scene image $I$ in the large vocabulary dataset LVIS, it is first fed into a backbone network $\phi(\cdot)$ (\eg, ResNet~\cite{ResNet}) to get the image feature $f\in \mathbb{R}^{d\times w\times h}$:
\begin{equation}
    f = \phi(I)\,,
\end{equation}
with dimensionality, width and height denoted as $d$, $w$, $h$, respectively. A region proposal network (RPN)~\cite{faster-rcnn} that contains both \emph{agnostic} classification and regression branches is then applied on the feature tensor $f$ to produce proposals $p$ from pre-defined anchor boxes. The ROIAlign~\cite{Mask-RCNN} then extracts proposal features:
\begin{align}
    p &= \text{RPN}(f) \,, \\
    f_p &= \text{ROIAlign}(f, p)\,,
\end{align}
where $p$ represents a large set of proposals, and $f_p$ is the set of aligned proposal features. $f_p$ goes into post-process modules (\eg, NMS~\cite{faster-rcnn}) before being sent to the RCNN head to get the final features set $f_n$:
\begin{equation}
    f_n = \text{RCNN}(f_p) \,,
\end{equation}
which is fed to classification and regression branches (\eg, linear layers) to produce the prediction results.

\subsection{Our three remedies for the regression bias}

\begin{figure*}
	\centering
		\includegraphics[width=0.95\linewidth]{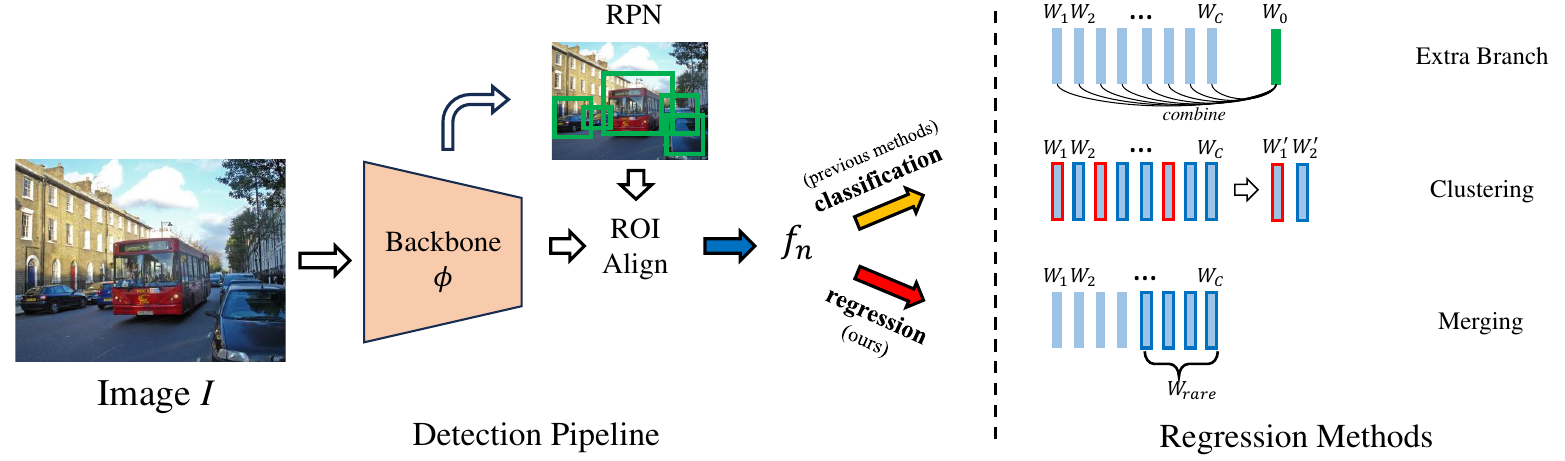}
	\caption{Illustration of the regular two-stage detection pipeline and the proposed regression methods. Previous methods (the left figure) generally focus on the final classification branch (the yellow arrow), while we focus on rectifying the regression bias (the red arrow). The right part shows our three regression methods, including adding an extra branch $W_0$, clustering regression heads (\eg, from $W_1,\dots,W_C$ to $W_1', W_2'$) and merging (\eg, merging rare categories into one regression head $W_{rare}$, \cf Table~\ref{tab:main-method}). In our main experiments, we choose `adding an extra branch' for its simplicity. This figure needs to be viewed in color.}
	\label{fig:method}
\end{figure*}

In Faster RCNN detection framework, there is a dedicated regression head for each class:
\begin{equation}
    r_i = W_i^T f_n ,\quad i=1,2,\dots,C\,,
\end{equation}
where $r_i = (\delta x_i, \delta y_i, \delta w_i, \delta h_i)$ represents the regression offset for class $i$, and $W_i$ is the class-specific regression head (a linear layer). We will now present our approaches to rectify the regression bias. The existing pipeline and our proposed methods are shown in Fig.~\ref{fig:method}.

\textbf{Extra class-agnostic branch.} This is a simple remedy to cope with the regression bias. Since rare class favor a class-agnostic head while class-specific ones are slightly preferable to frequent class in our hypothesis, we seek a tradeoff by using the combination of both heads. For class $i$, its regression head changes to:
\begin{equation}
    W_i' = \alpha  W_0 + (1-\alpha)  W_i \,,
\end{equation}
where $W_0$ is a shared class-agnostic regression head for \emph{all} classes, and $\alpha$ is a hyper-parameter to balance class-agnostic and class-specific heads. We empirically find that this simple change leads to consistent improvements over the default class-specific regression head (\cf Table~\ref{tab:main-method:extra}), while simply setting $\alpha = 0.5$ gives the optimal trade-off.

\begin{table*}
	\setlength{\tabcolsep}{2.5pt}
    \small
	\centering
    \subfloat[\textbf{Extra class-agnostic branch.} The weight ($\alpha$) ranges from 0 to 1.]
	{
		\label{tab:main-method:extra}
        \begin{minipage}{0.3\linewidth}
	        \centering
			\begin{tabular}{cllll}
				\toprule[1pt] 
				weight ($\alpha$) & AP & AP$_r$ &  AP$^{b}$ & AP$^{b}_{r}$  \\
				\midrule[1pt] 
				0.0 &  23.7 & 14.2 & 24.7 & 13.4 \\
				0.2 &  24.1 & 15.8 & 25.4 & 15.1 \\
				0.5 & \textbf{25.1} & \textbf{17.5} & \textbf{27.0} & 18.0  \\
				0.8 & 24.4 & 17.0 & 25.9 & 16.4 \\
				1.0 & 24.7 & 16.7 & 26.7 & \textbf{18.3} \\        
				\bottomrule[1pt]
			\end{tabular}            
        \end{minipage}
	}
    \hspace{1em}
    \subfloat[\textbf{Clustering heads.} Clustering using class size (`num') or class mean scale (`scale').]
	{
		\label{tab:main-method:cluster}
        \begin{minipage}{0.3\linewidth}
        \centering
		\begin{tabular}{cccllll}
			\toprule[1pt]
		    $k$ & num & scale  & AP &  AP$_r$  &AP$^{b}$ & AP$^{b}_{r}$ \\
			\midrule[1pt]
             base  & & & 23.7 & 14.2 & 24.7 & 13.4 \\        
             10 & \ding{51} &  & 24.5 & 14.5 & 26.3 & 14.4 \\
             10 & & \ding{51} & 24.2 & 12.8 & 26.0 & 12.6 \\
             100 & \ding{51} &  & 24.4 & 13.0 & 26.2 & 12.9 \\
             100 & & \ding{51} & \textbf{25.2} & \textbf{16.7} & \textbf{26.9} & \textbf{16.7} \\
			\bottomrule[1pt]
		\end{tabular}
    \end{minipage}
	}
    \hspace{1em}
    \subfloat[\textbf{Merging heads.} Directly merging regression heads in rare, common or frequent, respectively.]
	{
		\label{tab:main-method:merge}
        \begin{minipage}{0.3\linewidth}
        \centering
		\begin{tabular}{cllll}
			\toprule[1pt]
        	merge & AP &  AP$_r$ &  AP$^{b}$ & AP$^{b}_{r}$ \\
        	\midrule[1pt]
        	base & 23.7 & 14.2 &  24.7 & 13.4 \\
			$r$ &  25.1 & 16.3 & 26.7 & 16.7 \\
			$c$  & \textbf{25.5} & \textbf{17.7} & 27.2 & 17.2 \\
			$r,c$ & 25.3 & 17.1 & \textbf{27.3} & 18.2\\
			$r,c,f$  & 24.7 & 16.7 & 26.7 & \textbf{18.3} \\
        	\bottomrule[1pt]
		\end{tabular}
    \end{minipage}
}
\caption{Our three different methods to alleviate the regression bias. In~\ref{tab:main-method:extra}, we add an extra shared class-agnostic regression head during training and testing, where the shared class-agonistic head and the class-specific head are combined using a weight $\alpha$. In~\ref{tab:main-method:cluster}, we perform regression head clustering according to the number of of instances (`num') or mean box scale (`scale'). In~\ref{tab:main-method:merge}, we directly merge the regression heads of rare, common or frequent classes. The original class-specific heads in~\ref{tab:main-method:cluster} and~\ref{tab:main-method:merge} are replaced by the new heads. Note that `base' means the baseline method, i.e., with only class-specific regression heads. In~\ref{tab:main-method:extra}, $\alpha=0.0$ is equivalent to the baseline method, and $\alpha=1.0$ is equivalent to merging all $r,c,f$ heads into a single head (\ie, the last row in~\ref{tab:main-method:merge}).}
\label{tab:main-method}
\end{table*}

\textbf{Clustering heads.} This method is motivated by the analysis in Fig.~\ref{fig:motivation:scale}. Since some categories have similar statistics, we can assign them a \emph{shared} regression head to improve the generalization ability. We implement it by following three steps: sorting, grouping and assigning. First, we use the number of instance or mean box scale to sort the original categories in descending order ($C=1203$ in LVIS1.0):
\begin{equation}
    W_1,\dots,W_C \stackrel{sort}{\Longrightarrow}W_1^s, \dots, W_C^s \,.\\
\end{equation}
Then, we cluster them into $K$ groups. During clustering, we do not rely on time-cost algorithms such as K-means~\cite{DeepCluster} or GMM~\cite{GMM}), but simply put adjacent classes into one group, and each group has the same number of classes $N = \frac{C}{K}$:
\begin{equation}
\{(W_{N \times i+1}^s, W_{N \times i+2}^s, \dots, W_{N \times i+N}^s)\}_{i=0}^{K-1} \,,
\end{equation}
Finally, each group $i$ share one regression head:
\begin{equation}
    W_{gi} \stackrel{replace}{\Longleftarrow} (W_{N \times i+1}^s, W_{N \times i+2}^s, \dots, W_{N \times i+N}^s) \,.
\end{equation}
These shared regression matrix are then used for both training and testing. As shown in Table~\ref{tab:main-method:cluster}, clustering heads with similar scale statistics brings robust improvements over the baseline methods.

\textbf{Merging heads.} This approach has similar motivation to the previous one, which clusters regression heads, but is even more straightforward. We simply group the regression heads in pre-defined clusters. For example, we let all rare categories share a common regression head $W_{rare}$, and the same for common and frequent classes. Experimental results can be found in Table~\ref{tab:main-method:merge}, where we try four different combination (note that $r,c$ means merging rare and common classes all together into \emph{one} regression head). The results indicate that merging always leads to performance gains, especially for the rare categories. The most significant improvements come from merging only the common class (\ie, the row denoted as `$c$' in Table~\ref{tab:main-method:merge}).

The observation that merging common classes leads to the best improvement in AP$_r$ is somehow counter-intuitive. We conjecture that this is due to the \emph{partition shift} of rare, common and frequent in LVIS train and validation sets. In the LVIS1.0 train dataset, the size range for frequent, common and rare are [0, 404], [405, 865] and [866, 1202], respectively. While for validation set, they become [0, 212], [213, 536] and [537, 1202], respectively (\cf Fig.~\ref{fig:motivation:scale}). Hence, when we use training set statistics to merge the common class, a large portion of rare categories in the validation set are also potentially merged, thus contributing to the validation improvement on both AP$_r$ and AP$_c$.

\subsection{Picking one out of the three} 

All our proposed remedies improve the accuracy of long-tailed object detection and instance segmentation, which verifies the importance of rectifying the regression bias. We will choose \emph{adding a class-agnostic branch} (the first approach, \cf Table~\ref{tab:main-method:extra}) in our main experiments for the following reasons. On one hand, it does not need any dataset statistics, especially when the number of categories and the data distribution is unknown. On the other hand, by combining both class-agnostic and class-specific heads, it fully exploits object priors and per class knowledge: each type of head has its own merit, as Table~\ref{tab:motivation} shows.

Note that some results of \emph{merging heads} (not adopted in later experiments) are even better than the best accuracy in our chosen method, indicating that the results in our main experiment can be higher.

\section{Experiment}
\label{sec:exp}

We choose `adding an extra Class-Agnostic Branch' (abbreviated as CAB) to conduct our main experiments. We first combine CAB with existing long-tailed methods (\cf Table~\ref{tab:consistent-improve}), then choose `SeeSaw~\cite{LTD_SeeSaw} + CAB' as `Our' to compete with state-of-the-art methods (\cf Table~\ref{tab:compare-with-sota}). Although SeeSaw is a lower baseline than ECM~\cite{LTD_ECM}, it is more stable in reproducing (\cf the appendix). Finally, we generalize our methods to various evaluation metrics, different datasets and the mask branch. 

\subsection{Experimental settings}

\textbf{Datasets.} We use the large vocabulary dataset LVIS1.0~\cite{LVIS} as our main dataset, which contains 100k training and 20k validation images. Rare ($r$), common ($c$) and frequent ($f$) classes are defined by how many images they occur~\cite{LVIS}: [0, 10] for rare, [11, 100] for common, and (100, +$\infty$) for frequent, respectively. We also adopt COCO-LT~\cite{MS-COCO} and COCO2017~\cite{LTD_ClassificationFirst} to verify the generalization ability of our approach. COCO2017 is a large object detection dataset, containing 118k training and 5k validation images. It is relatively balanced in comparison with LVIS1.0. The COCO-LT dataset is an artificially sampled subset of COCO, with the same validation set but a long-tailed training set. It has about 99k training and 5k validation images. Following previous works~\cite{LTD_ClassificationFirst}, we partition COCO-LT into 4 evaluation subsets according to the number of training instances per class, with bins of [1, 20), [20, 400), [400, 8000) and [8000, -), respectively.

\textbf{Training details.} We reproduce four different methods as our baselines, including RFS~\cite{LVIS}, EQLv2~\cite{LTD_EQLv2}, SeeSaw~\cite{LTD_SeeSaw} and ECM~\cite{LTD_ECM}, following their default experiment settings. We employ MMDetection~\cite{mmdetection} as our detection framework to conduct our experiment, and train detection models of Faster-RCNN, Mask-RCNN and Cascade R-CNN for 1x or 2x scheduler (except Swin-Transformer based detectors, \cf the appendix), following previous works~\cite{LTD_PCB,LTD_ECM}. The batch size and learning rate are set as 16 and 0.02, and the data augmentation strictly follows previous long-tailed detection methods~\cite{LTD_ECM,LTD_EQLv2,LTD_SeeSaw}. During training, we use FP16 mixed precision training and the warmup strategy to stabilize the learning process. For the evaluation metrics, we adopt AP and AP$^b$ for instance segmentation and object detection, respectively, and adopt AP$_1^b$, AP$_2^b$, AP$_3^b$ and AP$_4^b$ on COCO-LT, corresponding to its 4 different subsets. With the suggested practice in LVIS official website, we run all our experiment 3 times on 8 RTX3090 GPUs to reduce the variance. Please refer to our supplementary material for more detailed information.

\subsection{LVIS detection and segmentation}

\textbf{Consistent improvements.} we first evaluate the effectiveness of our method on the LVIS1.0 dataset by combing the proposed approach `adding a class-agnostic branch' (CAB) with existing long-tailed object detection methods. Since our main focus is on bounding box regression, we list the object detection results in early columns and segmentation results in later ones. As shown in Table~\ref{tab:consistent-improve}, using CAB leads to consistent AP$^b$ and AP improvement over existing \emph{classification-based} methods, surpassing all of them with large margins. For object detection, Our CAB benefits the rare class a lot, with an increase of 4.6 AP$^b$ and 4.3 AP$^b$ on RFS and EQLv2. The same is true for instance segmentation, where a growing trend can be observed on all metrics, showing that CAB is also beneficial for later mask pixel predictions. Interestingly, the method `RFS+CAB' (which uses a CE loss) can almost achieve the same object detection accuracy as the SeeSaw method, and surpasses EQLv2 for about 1 AP$^b$. We thus conjecture that: besides merely focusing on classification, \emph{our regression methods} can serve as strong alternatives that also \emph{strongly boost} long-tailed detection accuracy.

\begin{table*}
	\small
	\centering
	\begin{tabular}{lcllllllll}
		\toprule[1pt]
		 \multirow{2}{*}{Method} & \multirow{2}{*}{+CAB} & \multicolumn{4}{c}{Detection} & \multicolumn{4}{c}{Segmentation} \\
         & &  AP$^b$  & AP$^{b}_{r}$ & AP$^{b}_{c}$ & AP$^{b}_{f}$ & AP & AP$_{r}$ & AP$_{c}$ & AP$_{f}$  \\
		\midrule[1pt]
        \multirow{2}{*}{RFS~\cite{LVIS}} & \textit{no} & 24.7 & 13.4 & 23.1 & 31.4 & 23.7 & 14.2 & 22.9 & 29.3 \\
         & \textit{yes} &  \textbf{27.0} & \textbf{18.0} & \textbf{25.3} & \textbf{32.9} & \textbf{25.1} & \textbf{17.5} & \textbf{23.9} & \textbf{29.7}\\
		\midrule

        \multirow{2}{*}{EQLv2~\cite{LTD_EQLv2}} & \textit{no} & 26.0 & 16.1 & 24.0 & 32.5 & 25.2 & 17.4 & 24.1 & 29.9\\
         & \textit{yes} & \textbf{28.1} & \textbf{20.4} & \textbf{26.3} & \textbf{33.5} & \textbf{26.0} & \textbf{19.5} & \textbf{24.9} & \textbf{30.2}\\
		\midrule
  
        \multirow{2}{*}{SeeSaw~\cite{LTD_SeeSaw}} & \textit{no} & 27.3 & 18.2 & 26.5 & 32.3  & 26.9 & 19.6 & 26.8 & 30.5\\
         & \textit{yes} &  \textbf{28.9} &\textbf{19.9} & \textbf{28.3} & \textbf{33.6} & \textbf{27.7} & \textbf{20.2} & \textbf{27.3} & \textbf{31.3}\\
        \midrule

        \multirow{2}{*}{ECM~\cite{LTD_ECM}} & \textit{no} & 27.7 & 17.7 & 26.6 & 33.1 & 27.2 & \textbf{19.6} & 26.6& 31.3\\
         & \textit{yes} &  \textbf{29.1} & \textbf{18.4} & \textbf{28.9} & \textbf{33.9} & \textbf{27.8} & 19.1 & \textbf{28.0} & \textbf{31.8}\\
        
		\bottomrule[1pt]
	\end{tabular}
\caption{Experiments on LVIS1.0. We combine four existing methods with our approach `adding a class-agnostic branch' (CAB). We reproduced RFS~\cite{LVIS}, EQLv2~\cite{LTD_EQLv2}, SeeSaw~\cite{LTD_SeeSaw} and ECM~\cite{LTD_ECM} using their official code. For clarity, here we list the object detection metrics AP$^b$ in the first four columns while putting the instance segmentation metrics AP in the later ones.}
\label{tab:consistent-improve}
\end{table*}

\textbf{Comparison with SOTA.} We then compare the proposed method with state-of-the-art methods using different object detection framework (Mask-RCNN, Cascade R-CNN) and backbones (ResNet-50, ResNet-101, Swin-T and Swin-B).  Note that `Our' means `SeeSaw + CAB'. For fair comparison, we reproduce majority of existing methods using their official released code unless specialized symbols (\eg, $^\dagger$) appears after a method's name. As can be seen in Table~\ref{tab:compare-with-sota}, our method achieves the overall highest accuracy in AP and AP$^b$. For ResNet series models, our regression technique easily surpass the best competitor ECM~\cite{LTD_ECM}, especially in AP$^b$ (an increase of 1.2 AP$^b$ for ResNet-50 and 1.4 AP$^b$ for ResNet-101). The advantage also holds for ViT-based object detectors, where we surpassed the best competitor ECM~\cite{LTD_ECM} with both Swin-Tiny and Swin-Base backbone architectures. Following LVIS's common practice, we didn't list AP$^b_r$ and AP$^b_c$ here, but we want to emphasize that the advantage of our regression methods can be further enlarged when more metrics are listed (\cf  Table~\ref{tab:consistent-improve}). This is also true if we replace `adding a Class-Agnostic Branch' (CAB) with the `merging heads' regression alternative (\cf the best accuracy in Table~\ref{tab:main-method}).

\begin{table*}
	\small
	\centering
	\begin{tabular}{llllllll}
		\toprule[1pt]
		 Architecture & Backbone  & Method & AP  & AP$_r$ & AP$_{c}$ & AP$_{f}$  & AP$^{b}$ \\
		\midrule[1pt]
        \multirow{8}{*}{Mask-RCNN~\cite{Mask-RCNN}} &\multirow{8}{*}{R50-FPN} & CE & 18.7 &  0.4 & 16.5 & 29.3 & 19.7\\
        & & RFS~\cite{LVIS} & 23.7 & 14.2 & 22.9 & 29.3 & 24.7 \\
        & & BSCE~\cite{BSCE} & 24.4 & 15.7 & 23.6 & 29.1 & 25.5\\
        & & EQLv2~\cite{LTD_EQLv2} & 25.2 & 17.4 & 24.1 & 29.9 & 26.0\\
        & & ECM~\cite{LTD_ECM}$^\dagger$ & 27.4 & 19.7 & 27.0 & 31.1 & 27.9\\
        & & ECM & 27.2 & 19.6 & 26.6 &31.3 & 27.7\\
        & & SeeSaw~\cite{LTD_SeeSaw} & 26.9 & 19.6 & 26.8 & 30.5 & 27.3\\
        & & Our & \textbf{27.7} & \textbf{20.2} & \textbf{27.3} & \textbf{31.3} & \textbf{28.9}\\

        \midrule
        
        \multirow{6}{*}{Mask-RCNN~\cite{Mask-RCNN}} &\multirow{6}{*}{R101-FPN} & CE~\cite{LTD_ECM}$^\dagger$ & 25.5 &  16.6 & 24.5 & 30.6 & 26.6\\
        & & EQLv2~\cite{LTD_EQLv2}$\dagger$ & 27.2 & 20.6 & 25.9 & 31.4 & 27.9\\
        & & ECM~\cite{LTD_ECM}$^\dagger$ & 28.7 & \textbf{21.9} & 27.9 & 32.3 & 29.4\\
        & & ECM & 28.6 & 20.9 & 28.4 & 32.2 & 29.3\\
        & & SeeSaw~\cite{LTD_SeeSaw} & 28.2 & 20.3 & 28.1 & 31.8 & 29.0\\
        & & Our & \textbf{29.0} & 21.0 & \textbf{28.9} & \textbf{32.4} & \textbf{30.7}\\

        \midrule
        
        \multirow{3}{*}{Cascade R-CNN~\cite{CascadeRCNN}} & \multirow{3}{*}{Swin-T~\cite{SwinTransformer}} & 
          ECM~\cite{LTD_ECM} & 34.1 & 23.7 & 34.9 & 38.0 & 37.6\\
        & & SeeSaw~\cite{LTD_SeeSaw} & 34.2 & 24.6 & 34.7 & 37.8 &  37.8 \\
        & & Our & \textbf{34.6} & \textbf{24.7} & \textbf{35.3} & \textbf{38.1} & \textbf{38.2}\\

        \midrule

        \multirow{2}{*}{Cascade R-CNN~\cite{CascadeRCNN}} & \multirow{2}{*}{Swin-B~\cite{SwinTransformer}} & 
          ECM~\cite{LTD_ECM}$\dagger$ & 39.7 & 33.5 & 40.6 & \textbf{41.4} & 43.6 \\
        & & Our & \textbf{39.9} & \textbf{34.5} & \textbf{40.7} & 41.1  & \textbf{44.2} \\
		\bottomrule[1pt]
	\end{tabular}
\caption{Comparison with state-of-the-art methods on the LVIS1.0 dataset. The results of `Our' means our proposed CAB regression method combined with the SeeSaw loss (`SeeSaw + CAB', \cf Sec.~\ref{sec:exp}). We show the metrics of instance segmentation AP and object detection AP$^b$. The results with a $^\dagger$ are copied from~\cite{LTD_ECM}, while others are reproduced by us using their official released code.}
\label{tab:compare-with-sota}
\end{table*}

\subsection{Generalization ability}

In this section, we will show the generalization ability of our regression methods in various aspects, including different evaluation metrics, datasets and its benefits on the mask branch.

\textbf{Different metrics.} We first explore how different metrics affect our model's accuracy. Two additional metrics are AP$_{boundary}$ (a more strict calculation method in instance segmentation~\cite{BoundaryIoU}) and AP$^{fixed}_{bbox}$ (constraining 10,000 predicted bounding boxes per class across the dataset~\cite{FixBox}). Object detector that obtain decent results in traditional metrics may not perform as well in these criterions. As shown in Table~\ref{tab:other-metrics}, our regression methods adapts well and surpasses all existing methods on both traditional and these challenging metrics. 

\begin{table}
	\setlength{\tabcolsep}{4pt}
	\small
	\centering
	\begin{tabular}{lcccc}
		\toprule[1pt]
		    Method & AP & AP$^b$ & AP$^{fixed}_{boundary}$ & AP$_{bbox}^{fixed}$\\
		\midrule[1pt]
         BAGS~\cite{LTD_BAGS} & 23.1 & 23.7 & - & 26.2 \\
         EQLv2~\cite{LTD_EQLv2} & 23.9 & 24.0 & 20.3 & 25.9\\
        SeeSaw~\cite{LTD_SeeSaw} & 25.2 & 25.4 &  19.8 & 26.5 \\
        ECM~\cite{LTD_ECM} & 26.3 & 26.7 & 21.4 & 27.4 \\
        \midrule
        Our & \textbf{27.2} & \textbf{28.0} & \textbf{22.1} & \textbf{28.2}\\
        \bottomrule[1pt]
	\end{tabular}
\caption{More evaluation metrics in LVIS1.0 dataset with a 1x scheduler using Mask RCNN. AP$_{boundary}^{fixed}$ and AP$_{bbox}^{fixed}$ are two newly proposed challenging metrics~\cite{BoundaryIoU,FixBox}.}
\label{tab:other-metrics}
\end{table}

\textbf{The COCO-LT dataset.} We also transfer our regression to another long-tailed dataset: COCO-LT. This is an artificially sampled~\cite{LTD_ClassificationFirst} subset of original COCO~\cite{MS-COCO}. We calculate the over bounding box metrics AP$^b$ and more fine-grained results: AP$_1^b$, AP$_2^b$ AP$_3^b$ and AP$_4^b$ (ranging from the rarest to the most frequent class). As shown in Table~\ref{tab:coco-lt}, our CAB leads to consistent improvements on all metrics (especially for the rarest categories AP$_1^b$) under different repeat-factor-sampling rates, showing the great power of CAB to help the rare classes. In fact, this conclusion generally holds true for all sampling rates in our experiments. We only listed three here for simplicity and clarity. 

\begin{table}
	\setlength{\tabcolsep}{4pt}
	\small
    \centering
	\begin{tabular}{lclllll}
		\toprule[1pt]
		Rate  & +CAB & AP$_{1}^b$  & AP$_{2}^b$ & AP$_{3}^b$ & AP$_{4}^b$& AP$^b$\\
		\midrule[1pt]
        \multirow{2}{*}{3e-3} & & 2.3 & 16.9 & 26.7 & 30.4 & 23.5  \\
         &  \ding{51} & \textbf{6.2} & \textbf{19.1} & \textbf{27.0} & \textbf{30.5} & \textbf{24.4}  \\ 
         \midrule
        \multirow{2}{*}{5e-3} & & 3.6 & 19.3 & 26.7 & 30.6 & 24.3\\
         &  \ding{51} & \textbf{7.5} & \textbf{20.0} & \textbf{27.3} & \textbf{30.7} & \textbf{25.0}\\
        \midrule 
        \multirow{2}{*}{1e-2} & & 8.7 & 20.3 & 27.8 & 30.3 & 
 25.2  \\
         &  \ding{51} & \textbf{10.8} & \textbf{21.6} & \textbf{28.0} & \textbf{30.5} & \textbf{25.8} \\ 
         \bottomrule[1pt]
	\end{tabular}
\caption{Experiments on COCO-LT~\cite{LTD_ClassificationFirst}. Following LVIS~\cite{LVIS}, we use a similar repeat-factor-sampling (RFS) strategy with different sampling rates (3e-3, 5e-3 and 1e-2).}
\label{tab:coco-lt}
\end{table}

\textbf{On balanced training set.} Furthermore, we validate how the proposed method perform on the relatively balanced dataset MS-COCO2017. We adopted Faster RCNN with three different backbones (ResNet-50, ResNet-101 and ResNext101-32x4d). Results in Table~\ref{tab:coco} clearly shows that CAB generalizes well to datasets with more balanced distributions. Interestingly, CAB boosts metrics of large and medium objects while shows similar accuracy on small objects. This is possibly because that small objects occupy over 60\% of the total instances while large objects only consumes roughly 15\%~\cite{SNIP}. Since our CAB is more beneficial to less-frequent classes, it brings higher performance gains in large objects than in small ones.

\begin{table}
	\setlength{\tabcolsep}{3pt}
	\small
    \centering
	\begin{tabular}{lcllllll}
		\toprule[1pt]
		  Backbone  & +CAB & AP  & AP$_{50}$ & AP$_{75}$ & AP$_{s}$  & AP$_{m}$ & AP$_{l}$ \\
		\midrule[1pt]
        \multirow{2}{*}{Res-50} & & 37.3 & 58.3 & 40.3 & 21.7 & 41.0 & 48.2 \\
         &  \ding{51} & \textbf{38.3} & \textbf{58.9} & \textbf{42.1} & \textbf{22.4} & \textbf{41.4}  & \textbf{50.4} \\ 
         
         \midrule
         
        \multirow{2}{*}{Res-101} & & 39.4 & 60.3 & 43.0 & 22.9 & 43.4 &51.0\\
         &  \ding{51} & \textbf{39.9} & \textbf{60.3} & \textbf{43.6} & \textbf{22.7} & \textbf{43.7}  & \textbf{52.8} \\ 

        \midrule
         
        \multirow{2}{*}{Res-32x4d} & & 41.0 & 62.2 & 44.6 & \textbf{23.9} & 45.3 & 52.9 \\
         &  \ding{51} & \textbf{41.7} & \textbf{62.5} & \textbf{45.4} & 23.6 & \textbf{45.9}  & \textbf{54.8} \\ 
        \bottomrule[1pt]

	\end{tabular}
\caption{Experiments on the (relatively) balanced COCO dataset. We adopted the Faster RCNN~\cite{faster-rcnn} detector and tried three different backbones (ResNet-50/101 and ResNext101-32x4d)}
\label{tab:coco}
\end{table}

\textbf{The mask branch.} Finally, we apply our CAB to the segmentation branch to test whether adding an class-agnostic prior is suitable for mask prediction. Experimental results are in Table~\ref{tab:mask-branch}, where we add a class-agnostic mask prediction head and combines it with each class-specific mask head. As shown in the table, our CAB generalizes well to segmentation tasks. Our main experimental results may be further improved if CAB was applied in both box and mask predictions.

\begin{table}
	\setlength{\tabcolsep}{3pt}
	\small
    \centering
	\begin{tabular}{ccllll}
		\toprule[1pt]
		  Branch  & +CAB & AP  & AP$_{r}$ & AP$^b$ & AP$_{r}^b$   \\
		\midrule[1pt]
         \multirow{2}{*}{Box} & & 23.7 & 14.2 & 24.7 & 13.4 \\
         & \ding{51} &  \textbf{25.1} & \textbf{17.5} & \textbf{27.0} & \textbf{18.0}  \\
         \midrule
         \multirow{2}{*}{Mask} & & 23.7 & 14.2 & 24.7 & 13.4 \\
         & \ding{51} & \textbf{24.1} & \textbf{15.0} & \textbf{25.0} & \textbf{13.9}  \\
        \bottomrule[1pt]
	\end{tabular}
\caption{Generalization to mask prediction. We added a class-agnostic mask prediction branch and combines it with each class-specific mask prediction results.}
\label{tab:mask-branch}
\end{table}

\textbf{Relations to the objectness branch.} We want to further clarify the relations (as well as differences) between our CAB and the objectness branch approach~\cite{LTD_ECM,LTD_SeeSaw} adopted in the \emph{classification} head (\ie, multiplying the objectness score to each class's prediction). We have argued that although they seemingly share similar structures with ours, they are essentially \emph{different} methods compared to ours, both conceptually and technically.

First, the purpose of objectness branch is to deal with the imbalanced distribution between foreground and background \emph{classification} samples~\cite{LTD_SeeSaw}, while our class-agnostic branch is motivated by the analysis that rare class much favor an agnostic \emph{regression} head, and we aim to tackle \emph{the regression bias}. Second, unlike classification where each class must preserve its own classifier, the regression heads can be clustered or merged, which provide more diverse solutions to reduce the regression bias besides CAB (\cf Table~\ref{tab:main-method}). Lastly, the effect of the objectness branch is yet to proved since it will \emph{decrease} the performance of rare and common classes even when combined with the pure CE loss~\cite{LTD_SeeSaw}, while our CAB (or clustering or merging heads) leads to consistent improvements, especially in rare class.

\subsection{Visualization and ablation}

In this section, visualization and ablation further illustrate the benefits of our regression methods.

\begin{figure*}
	\centering
		\begin{subfigure}{0.24\linewidth}
		\centering
			\includegraphics[width=0.49\linewidth, height=0.49\linewidth]{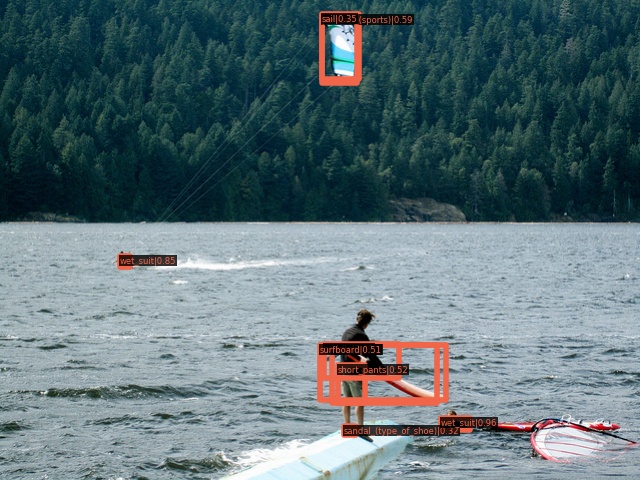}
			\includegraphics[width=0.49\linewidth,height=0.49\linewidth]{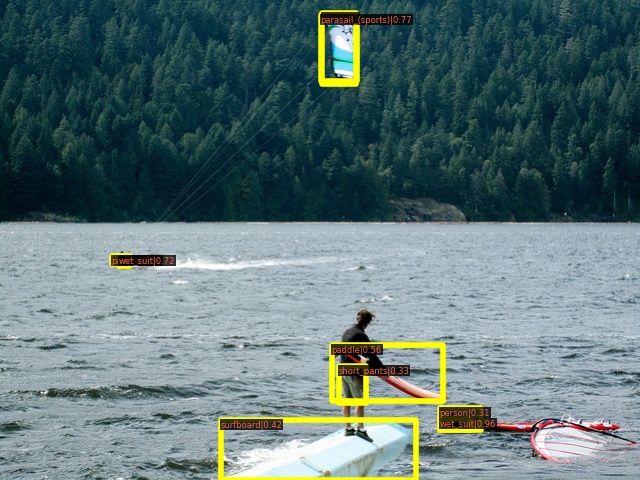}
		\end{subfigure}
		\begin{subfigure}{0.24\linewidth}
		\centering
			\includegraphics[width=0.49\linewidth, height=0.49\linewidth]{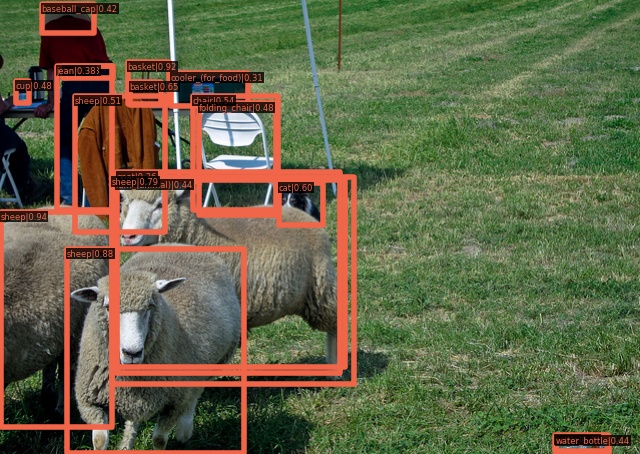}
			\includegraphics[width=0.49\linewidth,height=0.49\linewidth]{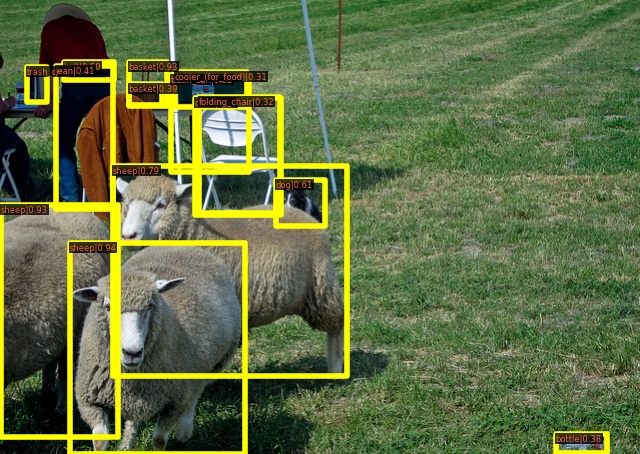}
		\end{subfigure}
		\begin{subfigure}{0.24\linewidth}
		\centering
			\includegraphics[width=0.49\linewidth, height=0.49\linewidth]{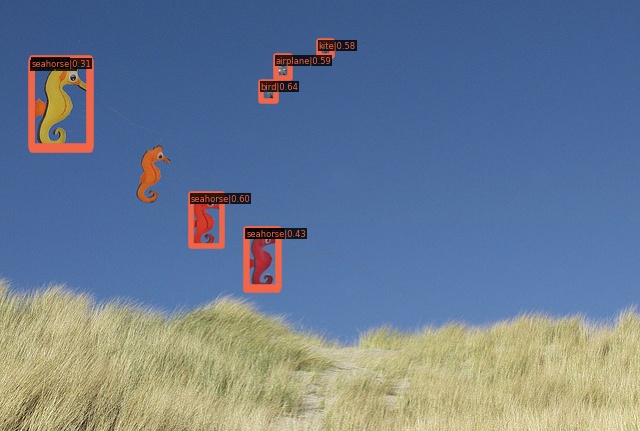}
			\includegraphics[width=0.49\linewidth,height=0.49\linewidth]{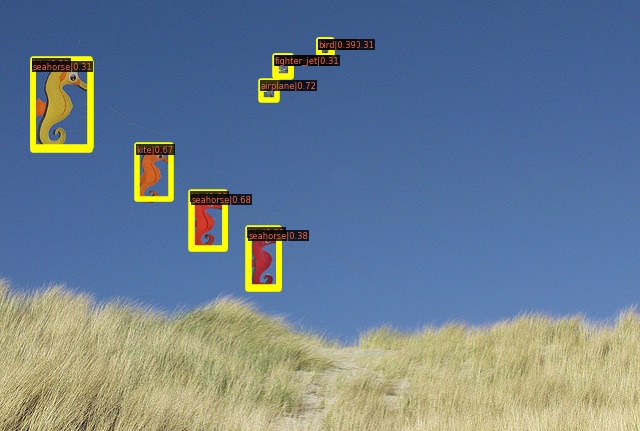}
		\end{subfigure}
		\begin{subfigure}{0.24\linewidth}
		\centering
			\includegraphics[width=0.49\linewidth, height=0.49\linewidth]{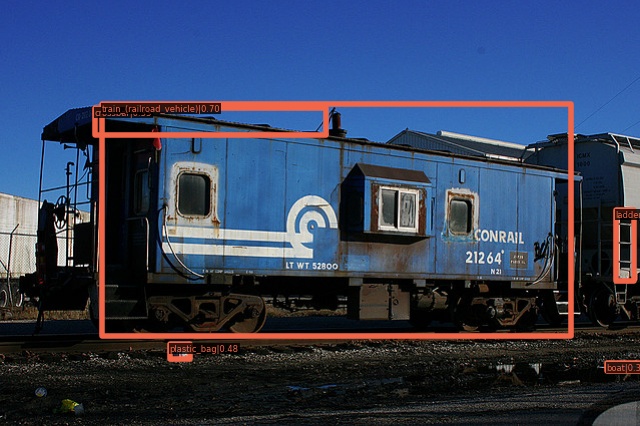}
			\includegraphics[width=0.49\linewidth,height=0.49\linewidth]{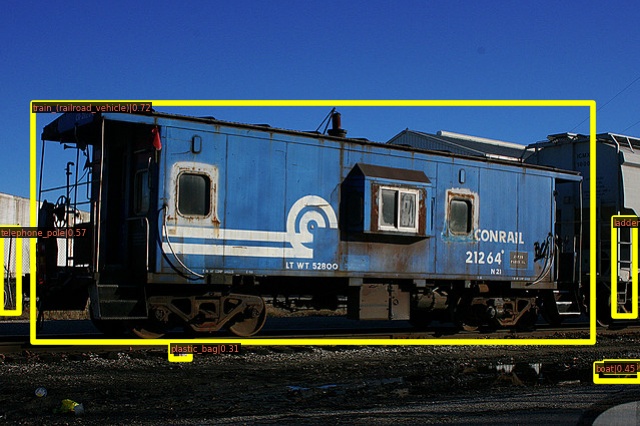}
		\end{subfigure}
	
		\begin{subfigure}{0.24\linewidth}
		\centering
			\includegraphics[width=0.49\linewidth, height=0.49\linewidth]{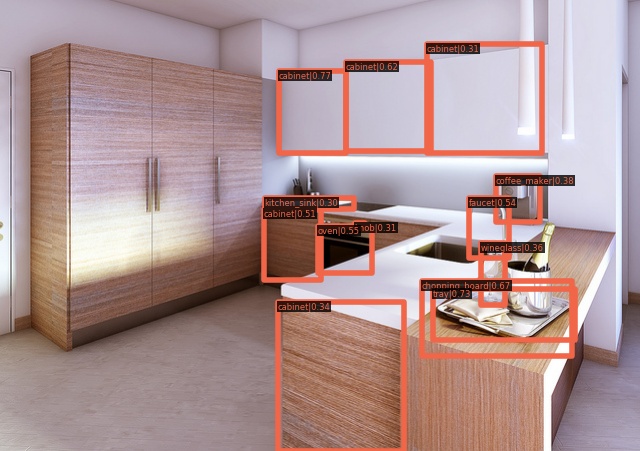}
			\includegraphics[width=0.49\linewidth,height=0.49\linewidth]{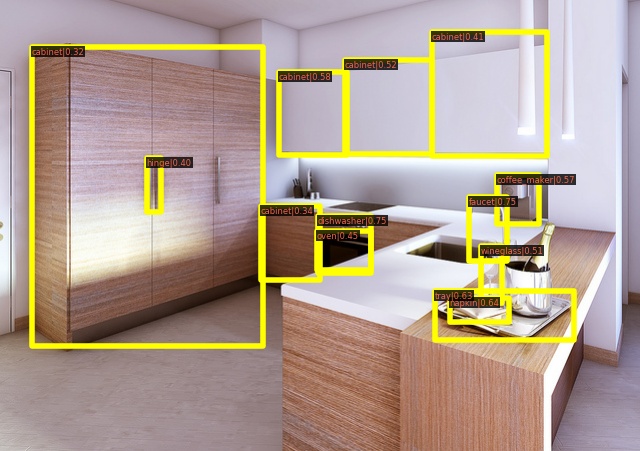}
		\end{subfigure}
		\begin{subfigure}{0.24\linewidth}
		\centering
			\includegraphics[width=0.49\linewidth, height=0.49\linewidth]{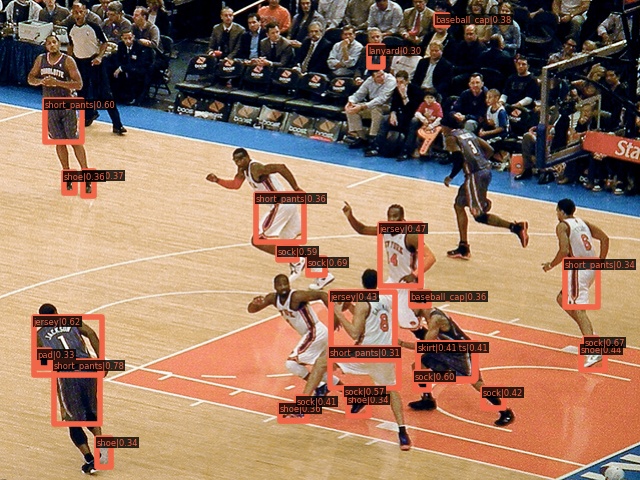}
			\includegraphics[width=0.49\linewidth,height=0.49\linewidth]{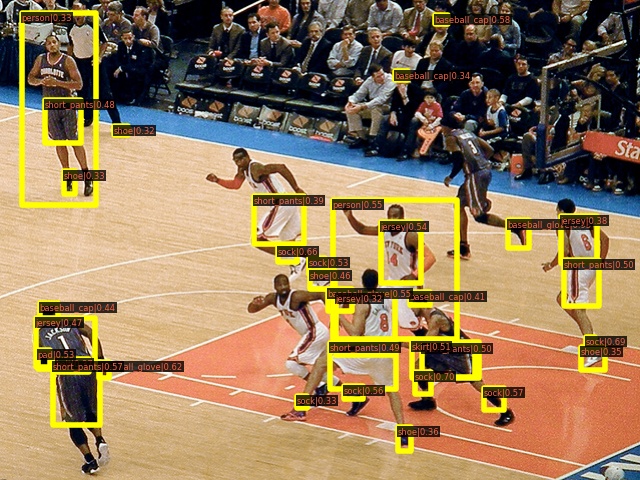}
		\end{subfigure}
		\begin{subfigure}{0.24\linewidth}
		\centering
			\includegraphics[width=0.49\linewidth, height=0.49\linewidth]{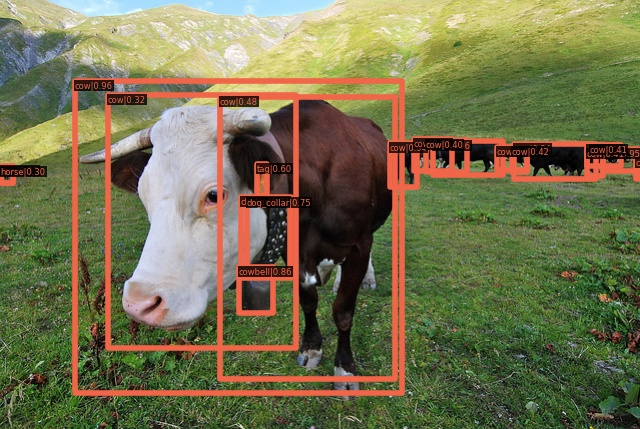}
			\includegraphics[width=0.49\linewidth,height=0.49\linewidth]{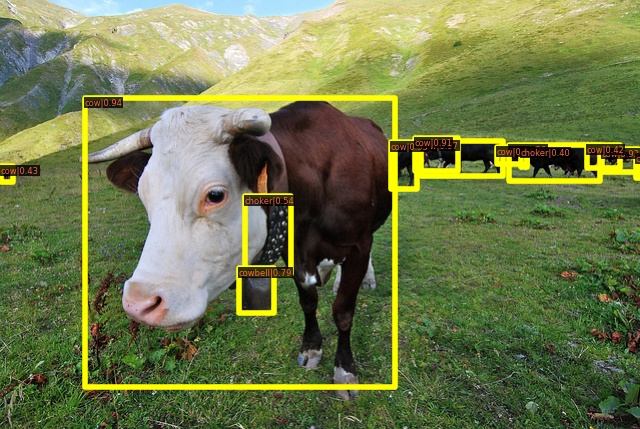}
		\end{subfigure}
		\begin{subfigure}{0.24\linewidth}
		\centering
			\includegraphics[width=0.49\linewidth, height=0.49\linewidth]{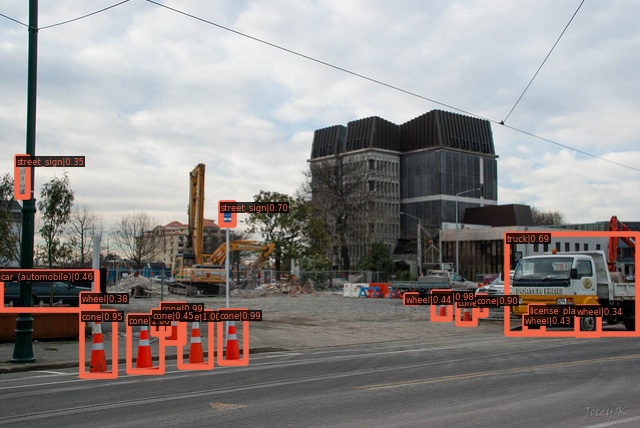}
			\includegraphics[width=0.49\linewidth,height=0.49\linewidth]{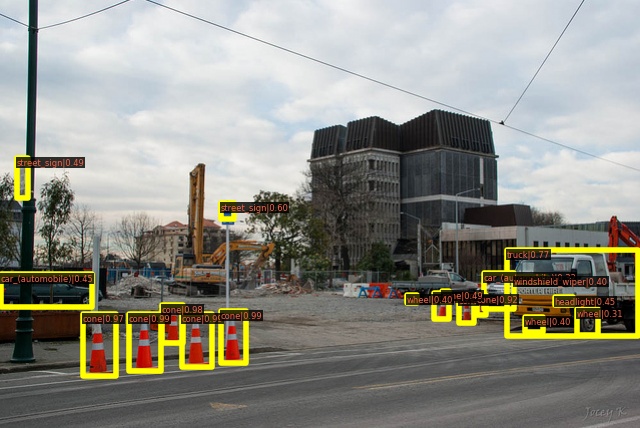}
		\end{subfigure}
		\caption{Visualizations of detection results before (in the left of each group) and after (in the right) using our CAB. We adopted RFS~\cite{LVIS} as the baseline in LVIS1.0 and combine it with our CAB regression method. In comparison, the proposed method is good at detecting missing objects, filtering duplicated objects away, as well as rectifying bounding box predictions. This figure needs to be viewed in color.}
\label{fig:box-refine}
\end{figure*}

\begin{figure}
	\centering
    \begin{subfigure}{0.49\linewidth}
    \centering
		\includegraphics[width=0.95\linewidth, height=0.7125\linewidth]{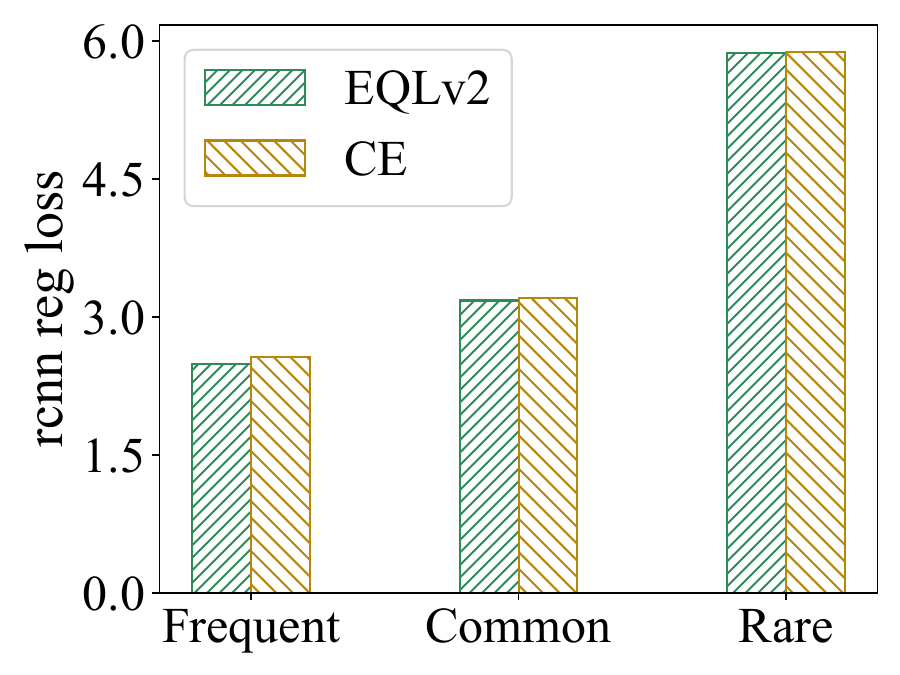}
    \caption{existing baseline methods}
    \end{subfigure}
    \begin{subfigure}{0.49\linewidth}
    \centering
		\includegraphics[width=0.95\linewidth,  height=0.7125\linewidth]{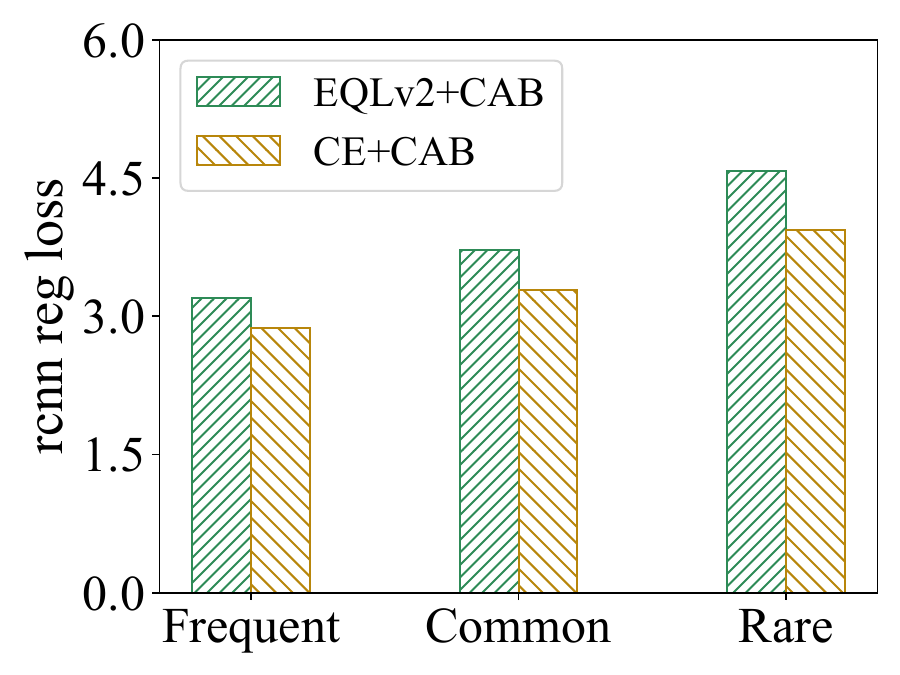}
    \caption{combined with our CAB}
    \end{subfigure}
    \caption{The loss distribution shift before and after combining with our CAB. Here we use EQLv2 and CE as baselines.}
	\label{fig:flatten-distribution}
\end{figure}

\textbf{Flatten distribution.} We first plot the RCNN regression loss of each category before and after combining with our regression CAB. The baseline methods we choose are EQLv2 and CE. As shown in Fig.~\ref{fig:flatten-distribution}, the regression loss for rare has seen a noticeable drop after adding our CAB, and the overall loss distribution has become much more balanced, too. We thus believe that the pipeline of our regression remedies can relieve the regression bias, further verifying that our hypothesis in Sec.~\ref{sec:intro} is indeed valid.

\begin{figure}
	\centering
	\begin{subfigure}{0.49\linewidth}
	\centering
		\includegraphics[width=0.95\linewidth, height=0.7125\linewidth]{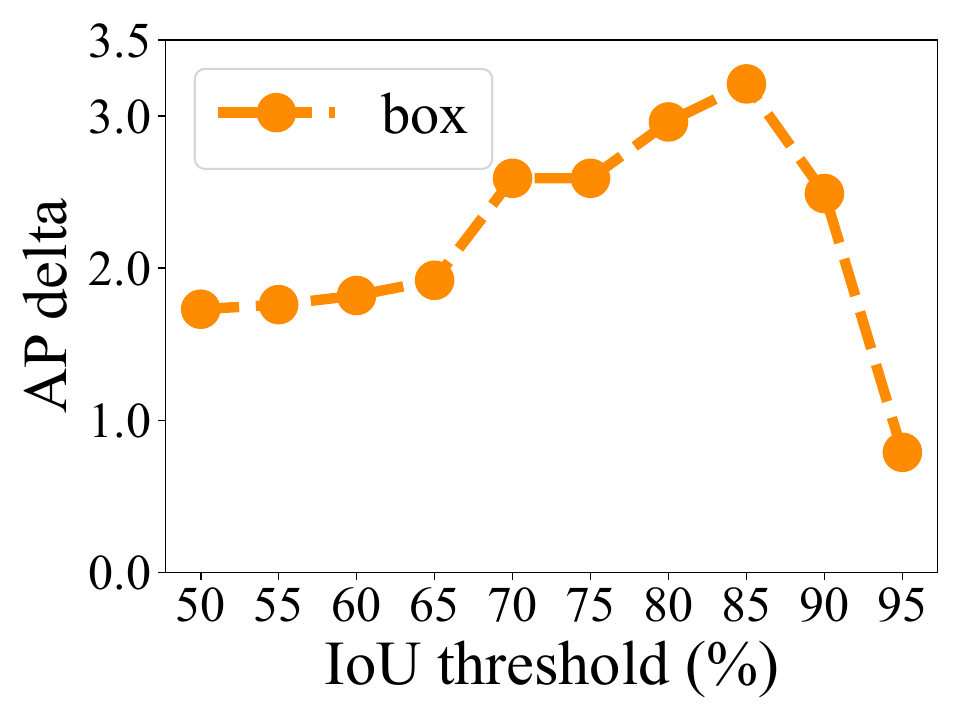}
	\caption{improvement on box}
	\label{fig:precise_ap:box}
	\end{subfigure}
	\begin{subfigure}{0.49\linewidth}
	\centering
		\includegraphics[width=0.95\linewidth,  height=0.7125\linewidth]{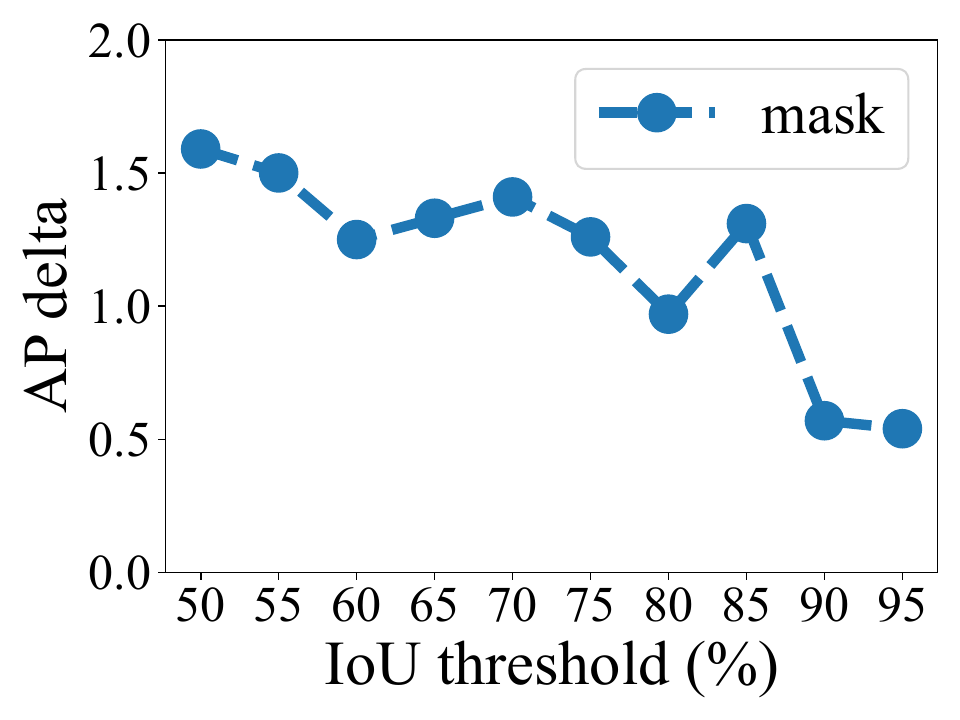}
	\caption{improvement on mask}
	\end{subfigure}
	\caption{The AP improvement of combining our CAB with the baseline RFS~\cite{LVIS} in LVIS1.0. We enumerate AP from AP50 to AP95, following the common practice adopted in MS-COCO.}
	\label{fig:precise_ap}
\end{figure}

\textbf{More accurate box/mask.} We then calculate AP50-AP95 of boxes and masks to find whether our CAB behaves well on stricter IoU thresholds. As show in Fig.~\ref{fig:precise_ap}, adding CAB achieves consistent accuracy gains over all IoU thresholds for both box and mask prediction, and is especially helpful for those hard IoU threshold (\eg, AP75-90) in box evaluation (\cf. Fig.~\ref{fig:precise_ap:box}). Since a higher threshold requires more precise box prediction, these results have shown that our regression methods is capable of producing precise boxes with better quality.

\textbf{Qualitative results.} Last but not least, we provide example detection images sampled from the LVIS1.0 validation dataset. For simplicity, we choose RFS~\cite{LVIS} with CE loss as baseline and combine it with our CAB method. As shown in Fig.~\ref{fig:box-refine}, both baseline and our CAB detect most of the objects in an image, but CAB generally caters for more details. For example, CAB can help discover missed boxes, like the cabinet in the lower left images. It is also clearly illustrated that CAB helps filter duplicate boxes away in the final predictions (\eg, the pictures with elephant and cow with grass background). Since CAB brings in better predictions, duplicate boxes will have larger overlap that may be suppressed by NMS (non-maximum suppression). If we zoom this figure (\cf the last colum in Fig.~\ref{fig:box-refine}), we will find that CAB is capable of rectifing the predicted boxes, and this can empirically explain why our CAB achieves much better AP under higher and more difficult IoU thresholds (\cf Fig.~\ref{fig:precise_ap:box}). 

\section{Conclusions and Limitations}

In this paper, we discovered that the regression bias (imbalanced regression loss distribution on the RCNN head) exists in long-tailed object detection, and adversely affects detection results. We thus proposed three remedies for rectifying the regression bias. The proposed method significantly boosts the performance of rare class AP$^b$, and achieves state-of-the-art results. We also generalize our regression methods to balanced dataset, different evaluation metrics and the mask branch. Finally, visualizations show that our method indeed produces better predicted bounding boxes.

As for the limitations, it remains unclear why the boosted accuracy become lower when the baseline are higher (\eg, the improvement of CAB on ECM is lower than that on RFS, \cf Table~\ref{tab:consistent-improve}). This may relate to the upper bound a backbone model can achieve. Since large network generally better fit the dataset (\cf Table~\ref{tab:compare-with-sota}), the performance a ResNet-50 can achieve is limited. We thus call for involving larger vision models on the LVIS dataset. Another limitations is the adaptability: the regression methods may not be easily applied to one-stage object detectors since they, unlike two-stage, usually only have class-agnostic regression head. However, almost all previous long-tailed object detectors also highly hinges on the two-stage structure (as pointed out by~\cite{LTD_EQFL}) to preserve its effectiveness, and it can be universally hard for long-tailed detector to be transferred to the one-stage pipeline. We will leave this as future work to fully excavate the potential of the regression branch.

{
\small
\bibliographystyle{ieeenat_fullname}
\bibliography{main}
}

\end{document}